\documentclass[runningheads,a4paper]{llncs}

\usepackage{amssymb}
\usepackage{graphicx}
\usepackage[T1]{fontenc}
\usepackage[latin9]{inputenc}
\usepackage[spanish,english]{babel}

\usepackage{float}
\floatstyle{plain}
\restylefloat{figure}
\floatstyle{plaintop}
\restylefloat{table}

\usepackage{url}
\urldef{\mailsa}\path|belanche@lsi.upc.edu|
\urldef{\mailsb}\path|jeronimo.hernandez@ehu.es|    
\newcommand{\keywords}[1]{\par\addvspace\baselineskip
\noindent\keywordname\enspace\ignorespaces#1}

\newcommand{\RR}[2]{{\mathbb #1}^{#2}}
  \newcommand{\R}{\RR{R}{}}
  
  \newcommand{\Rn}{\RR{R}{n}}
  
\newcommand{\HH}[2]{{\cal #1}^{#2}}
\newcommand{\st}{\hat{s}}

\newcommand{\supremo}{\sup\limits}
\newcommand{\infimo}{\inf\limits}
\newcommand{\smax}{s_{max}}

\newcommand{\sminL}{s_{0}}

\begin{document}

\mainmatter  % start of an individual contribution

\title{Similarity networks for classification:\\a case study in the Horse Colic problem}

\titlerunning{Similarity networks: a case study in the Horse Colic problem}

\author{Lluís Belanche\thanks{Corresponding author.}$^1$
\and Jerónimo Hernández$^2$}

\institute{$^1$Software Department, Computer Science School
 \\Technical University of Catalonia, Barcelona  (Spain)
 %, Jordi Girona, 1-3\\08034 Barcelona,
\\$^2$Intelligent Systems Group\\
 Department of Computer Science and Artificial Intelligence\\
 University of the Basque Country, Donostia (Spain)\\
% Manuel de Lardizabal, 1, 20018  \\
 \mailsa\\
 \mailsb}
 %\url{http://www.springer.com/lncs}} 

\maketitle

\begin{abstract}
  This paper develops a two-layer neural network in which the neuron
  model computes a user-defined {\em similarity function} between
  inputs and weights.  The neuron transfer function is formed by
  composition of an adapted logistic function with the mean of the
  partial input-weight similarities. The resulting neuron model is
  capable of dealing directly with variables of potentially different
  nature (continuous, fuzzy, ordinal, categorical). There is also
  provision for missing values. The network is trained using a
  two-stage procedure very similar to that used to train a radial basis
  function (RBF) neural network. The network is compared to two types of RBF networks in a non-trivial dataset: the Horse Colic problem, taken as a case study and analyzed in detail.
\end{abstract}


\section{Introduction}
\label{sec:Introduction}
 
The intuitive notion of \emph{similarity} is very useful to group
objects under specific criteria and has been used with great success
in several fields within or related to Artificial Intelligence, like
Case Based Reasoning \cite{OsborneBridge97}, Information Retrieval
\cite{BaezaYates99} or Pattern Matching
\cite{VeltkampHagedoorn00}. Under the conceptual cover of similarity,
we develop a class of neurons that accept heterogeneous inputs and
weights and compute a user-defined similarity function between these
inputs and weights. The neuron transfer function is the composition of
a parameterized sigmoid-like function adapted to the $[0,1]$ interval
taking the averaged vector of partial input-weight similarities as
argument. The basic idea is that a combination of similarity
functions, comparing variables independently, is more capable of
catching better the singularity of an heterogeneous dataset than other
methods which require \emph{a priori} data transformations. The
resulting neuron model then accepts mixtures of continuous and
discrete quantities, with explicit provision for missing
information. Other data types are possible by extension of the
model. The network is compared to two types of radial basis function
(RBF) networks in the Horse Colic dataset, which is analysed in detail
and used in two different classification tasks.  At least for one of
the tasks, the results point to a clear improvement in generalization
performance. An appealing advantage is found in the enhanced
interpretability of the learned weights, so often neglected in the
neural network community.

The paper is organized as follows. Section \ref{sec:Preliminaries}
further motivates the basis of the approach and reviews previous work
in relation to similarity measures and data heterogeneity; section
\ref{sec:ClusteringSimilarities} develops a clustering algorithm fully
based on similarity measures; section \ref{sec:hnn} builds the
similarity neural network. Finally, section \ref{sec:HCexpcomp}
presents experimental work.

\section{Preliminaries}
\label{sec:Preliminaries} 

\subsection{Data types and missingness}
%\label{sec:RelatedWork}

In many important domains from the real world, objects are described
by a mixture of continuous and discrete variables, usually containing
missing information and characterized by an underlying uncertainty or
imprecision. For example, in the well-known UCI repository \cite{UCI}
over half of the problems contain explicitly declared categorical
attributes, let alone other data types, usually unreported. In the
case of artificial neural networks (ANN), this {\em heterogeneous}
information has to be encoded in the form of real-valued quantities,
although in most cases there is enough domain knowledge to
characterize the nature of the variables.

The integration of heterogeneous data sources in information
processing systems has been advocated elsewhere \cite{Grecs}. In this
sense, a shortcoming of the existent neuron models is the difficulty
of adding prior knowledge to the model in a principled way.  Current
practice assumes that input vectors may be faithfully represented as a
point in $\Rn$, and the geometry of this space is meant to capture the
meaningful relations in input space. There is no particular reason why
this should be the case. Moreover, the activity of the units should
have a well defined meaning in terms of the input patterns
\cite{Omohundro}.  Without the aim of being exhaustive, commonly used
coding methods are (see, e.g. \cite{Proben}):

\begin{description}
\item [Ordinal] variables coded as real-valued or using a {\em thermometer} scale.
\item [Categorical] variables with $c$ modalities are coded using a
  binary expansion representation (also known as a 1-out-of-$c$ code).
\item [Vagueness] and uncertainty are considerations usually put aside.
\item [Missing] information is difficult to handle, specially when the
  lost parts are of significant size. Typical approaches remove the
  involved examples (or variables) or ``fill in the holes'' with the
  mean, median or nearest neighbor value. Statistical approaches need
  to model the input distribution itself \cite{Bishop95}, or are
  computationally very intensive \cite{missing}.

\end{description}

Although these encodings may be intuitive, their precise effect on network
performance (very specially in relation to overfitting) is far from clear.
This is due to the change in input distribution, the increase (sometimes
acute) in input dimension and other subtler effects, derived from
imposing an order or a continuum where there was none.

\subsection{Similarity measures}
\label{sec:SimilarityMeasures}

Let us represent patterns belonging to a space $X \neq \emptyset$ as a
vector $x$ of $n$ components, where each component $x_{k}$
represents the value of a particular feature $k$. A {\em similarity
  measure} is a unique number expressing how ``like'' two patterns
are, given these features. It can be defined as an upper bounded, exhaustive and total function $s : X \times X \rightarrow I_s \subset R$ with $|I_s|>1$ (therefore $I_s$ is upper bounded and $\smax \equiv \supremo_{\mathbb R} I_s$ exists). A similarity measure may fulfill many properties, like:

\medskip
\textbf{Reflexivity}: $s(x,y)= s_{max} \Leftrightarrow x = y$.

\smallskip
\textbf{Symmetry}: $s(x,y)= s(y,x)$. 

\smallskip
\textbf{Lower boundedness}: $\exists a \in R$ such that $s(x,y) \geq a$, for all $x,y \in X$ (note this is equivalent to ask that $\infimo_{\mathbb R} I_s$ exists). 

\smallskip
\textbf{Closedness}: given a lower bounded $s$, $\exists x,y \in X$ such that $s(x,y)= \infimo_{\mathbb R} I_s$ (equivalent to ask that $\infimo_{\mathbb R} I_s \in I_s$). 

\medskip
These axioms should be taken as {\em desiderata}. Some
similarity relations may fulfill part or all of them \cite{chan:pin}.
Other properties (like transitivity) may be of great interest in
some contexts, but are not relevant for this work. However, it is not difficult to show that reflexivity implies a basic form of transitivity \cite{Orozco04}.

\subsection{Similarity measures for different variable types}
\label{sec:hsi}

We present in this section specific similarity measures defined in a 
common co-domain $I_s=[0,1]$.  Not only it is possible to find
different types of variables, also different similarity measures could
be used for different variables of the same type. For notational
convenience, we use $s_{ijk}$ to mean $s_k(x_{ik}, x_{jk})$.

\vspace*{-4mm}
\subsubsection{Nominal variables}
It is assumed that no partial order exists among these values and the
only possible comparison is equality. The basic similarity measure for
these variables is the {\em overlap}. Let $x_{ik}, x_{jk}$ be the modalities taken by two examples $x_i, x_j$,
then $s_{ijk}  =  1$ if $x_{ik} = x_{jk}$ and 0 otherwise.

We introduce in this paper a frequency-based approach that goes beyond this simple equal/not-equal scheme:

\begin{eqnarray}
  s_{ijk} = \left\{
  \begin{array}{ll}
    0 & \mbox{ if } x_{ik} \neq x_{jk} \\
    1 - P_{ik} & \mbox{ if } x_{ik} = x_{jk} \\ 
  \end{array} \right.
\label{simfrequencybinbel}
\end{eqnarray} 

\noindent
where $P_{lk}$ is the \emph{fraction} of examples in a sample that take the value $x_{lk}$ for variable $k$
(ideally, one could use the \emph{probability} of this event, if this knowledge is available).
Therefore, if the values are different, there is not similarity. If they happen to be equal, then the similarity is inversely proportional to their probability. For instance, if two patients are being compared on their current illness, both having a rare illness makes them more similar than both having a very common one.
Other ways of inverting the probability are possible. In the absence of further knowledge, the linear one is the simplest choice.

\vspace*{-4mm}
\subsubsection{Ordinal variables}
\label{ord-cont}
These variables can be seen as a bridge between categorical and
continuous variables. It is assumed that the values of the variable
form a linearly ordered space $(\HH{O}{},\preceq)$. Let $x_{ik},
x_{jk} \in \HH{O}{}$, such that $x_{ik} \preceq x_{jk}$, and $P_{lk}$
be defined as above. Then,

\begin{equation}
s_{ijk} = \frac{2\log (P_{ik} + \ldots + P_{jk})}{\log P_{ik} + \log P_{jk}}
\label{simil:ordinal}
\end{equation}

where the summation runs through all the ordinal values $x_{lk}$ such
that $x_{ik} \preceq x_{lk}$ and $x_{lk} \preceq x_{jk}$ \cite{Lin98}.

\vspace*{-4mm}
\subsubsection{Continuous variables}
Let $x_{ik}, x_{jk} \in A=[r^-, r^+] \subset \R, r^+>r^-$. The standard
metric in $\R$ is a metric in $A$. Therefore, for any two values
$x_{ik},x_{jk} \in A$:

\begin{equation}
s_{ijk} = \st \left(\frac{|x_{ik}-x_{jk}|}{r^+ - r^-}\right)
\label{simil:continuous}
\end{equation}

where $\st: [0, 1] \longrightarrow [0,1]$ is a decreasing
continuous function. A very simple family is $\st(z) =
(1-z^\beta)^\alpha, 0 < \beta \leq 1, \alpha \geq 1$. We use here the
simplest choice $\alpha=\beta=1$.

\vspace*{-4mm}
\subsubsection{Fuzzy variables}
For variables representing fuzzy sets, similarity relations from the
point of view of fuzzy theory have been defined elsewhere
\cite{Zadeh78}, and different choices are possible. In possibility
theory, the {\em possibility} expresses the likeliness of
co-occurrence of two vague propositions, with a value of one standing
for absolute certainty. For two fuzzy sets ${\tilde A}, {\tilde B}$ of
a reference set $X$, it is defined as:

$$ \Pi_{(\tilde A)} ({\tilde B}) = \supremo_{u \in X} \ ( \mu_{\tilde A \,\cap
\tilde B}(u)) = \supremo_{u \in X} \ (\mbox{min}\ (\mu_{\tilde A}(u), \,
\mu_{\tilde B}(u)) ) $$

In our case, if ${\cal F}_k$ is an arbitrary family of
fuzzy sets, and $x_{ik}, \, x_{jk} \in {\cal F}_k$, the following
similarity relation can be used $s_{ijk} = \Pi_{(x_{ik})} (x_{jk})$.

%space*{-4mm}
%\subsubsection{Binary variables}
%In the data analysis literature there are many similarity measures
%defined on collections of binary variables (see e.g. \cite{Everitt}).
%This is mostly due to the uncertainty over how to accommodate negative
%(i.e. false-false) matches. The present situation is that of
%comparison of {\it a single} binary variable rather than a whole
%vector. In general, one should know which of the two matches is the
%really relevant (true-true or false-false). For these reasons,
%treating the variable as purely categorical can result in a loss of
%relevant information. Since this meta-knowledge is usually not handy,
%we use in this work a frequency-based approach, as follows.  Let
%$x_{ik}, x_{jk}$ be two binary values and let $P_{lk}$ be again the
%fraction of values of variable $k$ that take on the value $x_{lk}$.
%We define: $s_{ijk} = h(1-P_{ik}, 1-P_{jk})$, where $h(x,y) =
%\frac{2xy}{x+y}$ is the {\em harmonic mean} between $x$ and $y$. This
%measure compares the {\em agreement on the rarity} of each value: the
%similarity is higher the less frequent the values are.

%\footnote{The difference between a two-valued categorical variable
%  and a binary variable is that in the latter the two values are
%  mutually exclusive. For instance, the variable {\it color} with
%  values {\it black, white} would be categorical, whereas the variable
%  {\it is-black?} with values {\it yes, no} would be binary.
%  Sometimes the distinction is more conceptual than practical.}

\subsection{Missing value treatment}
\label{sec:MissingInformation}

Missing information is a recurrent problem in data analysis because
there are many causes for the absence of a value. The problem acquires
more relevance when significant parts of a data sample are lost or
unknown. There are basically three ways of dealing with \emph{missing values}:
fill in the examples, extend the learning methods to cope with
incomplete data or discard the examples (or the variables) with missing values.
We advocate for the second possibility, for which there exist some possible approaches:

\begin{enumerate}
\item The first proposal is based on Gower's general similarity
  measure \cite{Gower71}: 

\begin{equation}
S_G(\vec{x}_i,\vec{x}_j) = \frac{\sum\limits_{k=1}^n s_{ijk} \delta_{ijk}}{\sum\limits_{k=1}^n \delta_{ijk}}
\end{equation}
\noindent

where $s_k$ is a partial similarity function to be aggregated and
$\delta_{ijk} \in \left\lbrace 0 , 1 \right\rbrace$ is equal to 0 every
time $s_{ijk}$ is \emph{missing} (because one or both of $x_{ik},
x_{jk}$ are missing). It is not difficult to realize that this is
equivalent to the replacement of the missing similarities by the
average of the non-missing ones. Therefore, the conjecture is that the
missing values, if known, would not change the overall similarity.

\item The second proposal is even simpler: to replace the missing
  partial similarity measures by a constant quantity, namely the
  midpoint of the similarity co-domain $I_s$. For instance, when
  $I_s=[0,1]$, this constant would be $\frac{1}{2}$.  Doing this we
  are assuming that the missing similarities, if known, would make the
  example as similar to any other example as the average similarity.
\end{enumerate}

Both methods look very naive and indeed they are; on the other hand, they are very intuitive and computationally simple.
The appealing trait of these two approaches is that they do not try to estimate the missing information (a delicate and risky undertaking) but to estimate the \emph{overall} similarity between two observations, given that some of the partial similarities could not be computed. We argue that this second task is easier and, after all, is what we really are interested in: the \emph{similarity value}. This is the reason why we consider missing value treatment together with the construction of the overall similarity value.

\subsection{Normalized aggregation of similarities}
When we aggregate (e.g. by averaging) the partial similarities we are
assuming that all of them have the same importance. However, each
partial similarity covers its co-domain $[0,1]$ in a different way. The
partial similarities that accumulate on the upper half of the interval
have more influence in the overall value, because they do a more
important contribution to the aggregation.  We argue that this biased
behavior should be corrected so that all the partial similarities have
a common baseline.

Let $s_{..k}$ be the mean similarity among all examples in the
analyzed data sample, according to variable $k$ only.  We first
rescale all the similarities as $\hat{s}_{ijk} =
\frac{s_{ijk}}{s_{..k}}$. Then a normalization function $n:
(0,+\infty) \rightarrow (0, 1)$ is applied:

\begin{equation}
  n(z) = \frac{z^a}{z^a+1}
\label{normalize}
\end{equation}

where $a$ conveniently controls the shape of the function.  When a
similarity computation is needed, it is calculated as
$n(\hat{s}_{ijk})$ instead of $s_{ijk}$.  The decision on missing
values in section \ref{sec:MissingInformation} can now be better
justified. The similarity between two elements $x_{ik},x_{jk}$ is
now computed as:

\begin{eqnarray}
  s_{ijk} = \left\{
  \begin{array}{ll}
    n\left(\frac{s(x_{ik},x_{jk})}{s_{..k}}\right) & \mbox{ if neither of $x_{ik}, x_{jk}$ are missing} \\
    \frac{1}{2} & \mbox{ otherwise} \\
  \end{array} \right.
\label{simmissing}
\end{eqnarray} 

This is so because, when $s_{..k}$ is used to replace the missing
similarity \emph{value}, we have
$n(\frac{s(x_{ik},x_{jk})}{s_{..k}})=n(\frac{s_{..k}}{s_{..k}}) =
\frac{1}{2}$ (this holds regardless of the value of $a$).

%The effect is that the new similarity values are spread out 

\section{Clustering similarity data}
\label{sec:ClusteringSimilarities}

In a clustering task the examples are grouped attending to some
similarity measure.  The \textsc{Leader} algorithm is a simple and
attractive unsupervised clustering method \cite{Hartigan75}.  In
essence, the algorithm processes the examples of the dataset taking
one at a time and evaluating if it can belong to any cluster already
created. If it cannot, a new cluster will be created using this new
example as leader.

We have developed a new \textsc{Leader 2} version of the algorithm
that represents an improvement in two ways. First, the algorithm now
works using general similarities instead of metric distance
functions. Second, given $\sminL$, the method is guaranteed to fulfill
a number of interesting properties:

\begin{enumerate}
  \item For any example, the similarity with its leader is \emph{at least} $\sminL$.
  \item The similarity between any two leaders is \emph{lower} than $\sminL$.
  \item If two examples are repeated in the dataset, they will have the \emph{same} leader.
  \item For any example, the similarity with its leader is \emph{higher} than that with
    any other leader.
\end{enumerate}

One immediate consequence of these properties is that the lowest
similarity of an example with its leader will be higher than the
highest similarity between two different leaders.

In summary, the algorithm needs the specification of one parameter
($\sminL \in I_s$) and the returned leaders are a subset of
the data set (thus there is no problem in delivering ``impossible
centroids'' as many algorithms do). The number of clusters cannot be
estimated beforehand, but it is possible to establish a relationship
with the $\sminL$ parameter: a higher $\sminL$ implying a higher
number of clusters.

\section{Similarity Neural Networks} 
\label{sec:hnn}

\subsection{The S-Neuron Model}

Consider $\,s : {\cal H}^{n} \times {\cal H}^{n} \rightarrow I_s$ a
similarity function in ${\cal H}^{n}={\cal H}^{(1)} \times\ldots\times
{\cal H}^{(n)}$, the cartesian product of an arbitrary number $n$ of
{\it source sets}, where $I_s=[0,1]$. This function is formed by
combination of $n$ partial similarities $\,s_k : {\cal H}^{(k)} \times
{\cal H}^{(k)} \rightarrow I_s$, $k=1,\ldots,n$, each ${\cal H}^{(k)}$
being the domain of the predictive variable $k$.
  
The $s_k$ are normalized to a common real interval ($[0,1]$ in this
case) and computed according to different formulas for different
variables (possibly but not necessarily determined by variable type
alone). A {\em neuron model} can be devised that is both
similarity-based and handles data heterogeneity and missing values, as
follows. Let $\Sigma_i(\vec{x})$ the function computed by the $i$-th
neuron, where $\vec{x} \in {\cal H}^{n}$ having a weight vector
${\mu}_i \in {\cal H}^{n}$ and smoothing parameter $p_i$, defined as:

\begin{eqnarray}
  \Sigma_i(\vec{x}) = f(s(\vec{x},\mu_i), p_i),\ \mbox{ with } s(\vec{x},\mu_i) = \frac{1}{n}\sum_{k=1}^n s_k(x_k,\mu_{ik})
\label{functsneuron}
\end{eqnarray} 

\begin{figure}[!htp]
\begin{center}
\includegraphics[scale=0.8]{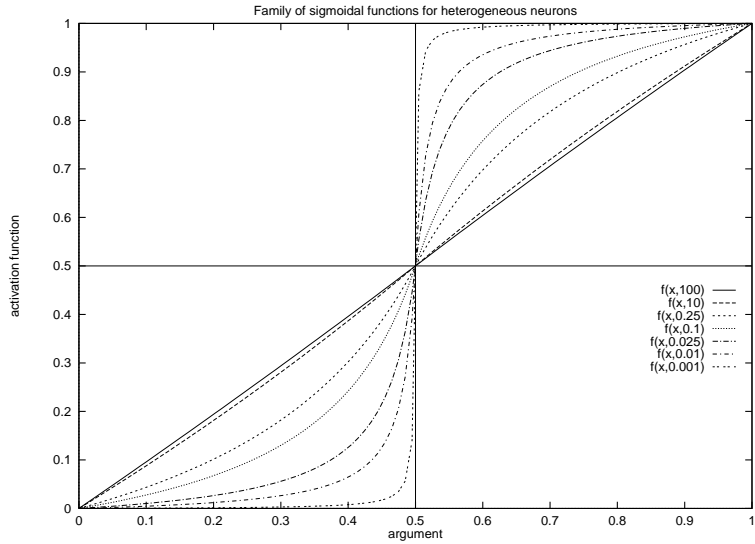}
\caption{Family of sigmoidal functions $f(\cdot, p)$ for different values of $p$.}
\label{figFp}
\end{center}
\end{figure}

This S-neuron adds a non-linear {\em activation} function to the
linearly aggregated similarities.  Such function could be any
sigmoid-like automorphism (a monotonic bijection) in $[0,1]$. In
particular, we consider the simple family of functions:

\begin{eqnarray}
  f(x, p)= \left\{
  \begin{array}{ll}
    \frac{-p}{(x-0.5)-a(p)} -a(p) & \mbox{ if } x \leq 0.5 \\
    \frac{-p}{(x-0.5)+a(p)} +a(p)+ 1 & \mbox{ if } x \geq 0.5 \\ 
  \end{array} \right. \nonumber \\
  a(p) = \frac{-0.5 + \sqrt{{0.5}^2 + 4p}}{2}
\label{functFp}
\end{eqnarray}

\noindent
where $p>0$ is a parameter controlling the shape of the function
(Fig. \ref{figFp}).  The function fulfills $\forall p \in \RR{R}{+},\
f(0,p)=0,\ f(1,p)=1,\ \lim\limits_{p \to \infty} f(x,p)=x$ and
$f(x,0)=H(x-0.5)$, being $H$ the Heaviside function.

\subsection{Similarity Neural Networks}
Similarity neural networks (SNN) are neural architectures built out of
the previously defined $S$-neurons, thus allowing for heterogeneous
or missing inputs. A feed-forward architecture, with a hidden
layer composed of heterogeneous neurons and a linear output layer
is a straightforward choice, thus conforming a {\em hybrid} structure.
The $k$-{th} output neuron of the SNN computes the function:

\[
f_k(\vec{x}) = \sum_{i=1}^{h} w_{ki}\Sigma_i(\vec{x})+w_{k0},\ k=1,\ldots,m
\]

where $h>0$ is the number of hidden $S$-neurons, $m$ is the number of
outputs and $\{w_{ki}\}$ is the set of mixing coefficients. The SNN
thus keeps linearity in the output layer and interpretability in the
hidden layer. It can be naturally seen as a generalization of the
RBF. This is so because the response of hidden neurons is localized:
centered at a given object (the neuron weight, where response is
maximum), falling down as the input is less and less similar to this
center.

\subsection{Training the \textsc{SNN}}
Let $\{(\vec{x}_l,y_l)\}_{l=1}^N$ represent a training data sample.
Since the \textsc{SNN} is a two-layer feed-forward neural network with
local computation units in the first layer, training can be solved
very efficiently in a two-stage procedure, as detailed next:

\subsubsection{First layer weights}
The first layer centers are a subset of the examples in the sample
dataset. These centers are chosen to be the cluster \emph{leaders}
returned by the \textsc{Leader 2} clustering algorithm acting on the
set $\{\vec{x}_l\}_{l=1}^N$ (hence $h$ is set to the number of
leaders). This algorithm uses the user-defined similarity as explained
in previous sections.

\subsubsection{The value of $p$}
Based on the information delivered by the clustering, we associate the
\emph{compactness} of a cluster with a greater slope of the $f$
function (a lower $p$, Fig. \ref{figFp}). When a cluster is more
compact (there is a big number of examples with a high similarity with
the leader), it is easier to decide whether a new example belongs to
that cluster or not because the cluster is well-defined and the limits
are clear.  This situation corresponds to $f(\cdot,p)$ working similar
to a Heaviside function ($p \rightarrow 0$).  This behavior can be
achieved by computing first a \emph{relative compactness index}:
\begin{equation}
  \chi_i = \frac{m_i l_i}{m_i l_i + \alpha \overline{m} \overline{l}}
\end{equation}
\noindent

where $l_i$ is the number of examples in cluster $i$ and $m_i$ is the
average similarity of these examples to their leader; the quantities
$\overline{m}$ and $\overline{l}$ are the corresponding global
averages (across the whole clustering). This index can be used to
obtain the smoothing parameter required in eq. \ref{functsneuron} as
$p_i=-\ln \chi_i$. The value of $\alpha$ is set at $\exp(0.1) - 1$,
which is related to the value of $p=0.1$ corresponding to an
``average'' compact cluster (relative to the current clustering).

\subsubsection{Second layer weights}
Regularization is a technique that incorporates additional information
to the fit, usually a complexity penalty to prevent overfitting.
  
\[
SSE_\lambda = \sum_{i=1}^{N} \sum_{k=1}^{m} \left(y_{ki} - f_k(\vec{x}_i)\right)^2 + 
      \lambda\sum_{j=1}^{h} \sum_{k=1}^{m} w_{kj}^2
\]
\noindent

where the first term is the sum of squared errors and the second is
the regularization term (known as \emph{ridge regression}, in this
case). The minimization of $SSE_\lambda$ forces to compensate smaller
errors against smaller weights. We define the $H=(h_{ij})$ matrix as
$h_{ij}=\Sigma_j(\vec{x}_i), i=1,\ldots,N, j=0,\ldots,h$.

\begin{eqnarray}
  H = \left[
  \begin{array}{ccccc}
    \Sigma_0(\vec{x}_1) & \Sigma_1(\vec{x}_1) & \Sigma_2(\vec{x}_1) & \dots & \Sigma_{h}(\vec{x}_1) \\
    \Sigma_0(\vec{x}_2) & \Sigma_1(\vec{x}_2) & \Sigma_2(\vec{x}_2) & \dots & \Sigma_{h}(\vec{x}_2) \\
    \ldots & \dots & \dots & \ddots & \vdots \\
    \Sigma_0(\vec{x}_{N}) & \Sigma_1(\vec{x}_{N}) & \Sigma_2(\vec{x}_{N}) & \dots & \Sigma_{h}(\vec{x}_{N})\\
  \end{array}
  \right] 
\end{eqnarray} 

\noindent
where $\Sigma_0(\cdot)=1$. Let $A = H^TH + \lambda I$, $P =
I-HA^{-1}H^T$, and $y$ the vector of outputs, where $I$ is an identity
matrix of appropriate dimensions. The optimal weight vector is
$\vec{w}^*=A^{-1}H^Ty$, the minimizer of $SSE_\lambda$ for a certain
$\lambda$. The \emph{Generalized Cross Validation} error is:

\begin{equation}
  {GCV} = \frac{N y^TP^2y }{\left(Tr\left( P \right)\right)^2 }
\label{functcriteriagcv}
\end{equation}
\noindent

When the derivative of $GCV$ is set to zero, the resulting equation can be manipulated so that one $\lambda$ appears isolated in one side of the equation. The value of $\lambda$ can be re-estimated iteratively until convergence \cite{Orr96}, using

\begin{equation}
  \lambda = \frac{Tr\left( A^{-1} - \lambda A^{-2} \right) \cdot y^T P^{2} y}
                 {Tr\left( P \right) \cdot (\vec{w}^*)^TA^{-1}\vec{w}^*}
\label{functlambda}
\end{equation}

An initial guess for $\lambda$ is used to evaluate the updating
expression, which produces a new guess. The obtained sequence
converges to a local minimum of GCV. In this work, this initial set is
$\lambda \in \{10^{-6}, 10^{-3}, 1\}$. In addition, a maximum number
of 100 iterations is set.

\section{A case study: the Horse Colic problem}
\label{sec:HCexpcomp}

We develop in this section a fully worked application example on a
challenging dataset.  This problem has been selected as characteristic
of modern modeling problems because of the diversity in data
heterogeneity and the presence of missing values
\cite{XuWunsch08}. This dataset is made available at the UCI
repository \cite{UCI} and created by M. McLeish and M. Cecile
(Computer Science Department, University of Guelph, Ontario,
Canada). Each example is the clinical record of a horse. The
attributes (variables) are specially well documented.  The number of
examples is modest, and therefore the chances of overfitting are
increased due to a bad pre-processing. In summary, there are 368
examples described by 28 attributes (continuous, discrete, and
categorical) and a 30\% of missing values.

\subsection{The Horse Colic dataset}

The problem description and the dataset itself are taken from the UCI
repository \cite{UCI}. The available documentation is analyzed for an
assessment on the more appropriate treatment. Missing information is
also properly identified. There are several possible tasks that can be
chosen for this dataset. The two most common settings are the
prediction of attributes 23 ('what happened to the horse?') and 24
('was the problem (lesion) surgical?'), using attributes 1,2 and 4 to
22 as predictors. We call these two separate tasks \textsc{HC23} and
\textsc{HC24}, respectively.

In task \textsc{HC23} there are two examples less because these two
examples have a missing value in the class variable. In
accordance to the documentation, attributes 3 and 28 are not used
because they do not provide useful information. Attributes 25, 26 and
27 ('type of lesion?') are also discarded because they represent
alternative class variables.  It should be noted that the missing
value counts are based on the full dataset.  Class distribution is as
follows:

\begin{description}
	\item[HC23]: what eventually happened to the horse (lived
- 61.5\%, died - 24.3\% or was euthanized - 14.2\%)
  \item[HC24]: was the lesion surgical? (yes - 63\% or no - 37\%). 
\end{description}

The following list details the used attributes, their characteristics and the decision taken on the type of attribute.

%The main characteristics are displayed in Table~1. 

{\small
\begin{itemize}
  \item Variable 1: \textbf{Surgery?} (Yes, it had surgery; It was treated without surgery)\\
    Comments: none.\\
    Decision: \texttt{Categorical}.

  \item Variable 2: \textbf{Age} (Adult horse; Young (< 6 months))\\
    Comments: none.\\
    Decision: \texttt{Categorical}.

 \item Variable 3: \textbf{Hospital number}\\
    Comments: the case number assigned to the horse.\\
    Decision: REMOVED.

  \item Variable 4: \textbf{Rectal temperature}\\
    Comments: Temperature of the horse in degrees Celsius.\\
    Decision: \texttt{Continuous}.

  \item Variable 5: \textbf{Pulse}\\
    Comments: The heart rate in beats per minute.\\
    Decision: \texttt{Continuous}.

  \item Variable 6: \textbf{Respiratory rate}\\
    Comments: none.\\
    Decision: \texttt{Continuous}.

  \item Variable 7: \textbf{Temperature of extremities} (Normal; Warm; Cool; Cold)\\
    Comments: an indicator of the peripheral circulation. 
    The values are re-ordered as: cold, cool, normal, warm.\\
    Decision: \texttt{Ordinal}.

  \item Variable 8: \textbf{Peripheral pulse} (Normal; Increased; Reduced; Absent)\\
    Comments: The values are re-ordered as: Absent; Reduced; Normal; Increased.\\
    Decision: \texttt{Ordinal}.
    
  \item Variable 9: \textbf{Mucous membranes} (normal pink; bright pink; pale pink; pale cyanotic; bright red / injected; dark cyanotic)\\ 
    Comments: a measurement of membrane color. Could it be considered ordinal?\\
    Decision: \texttt{Categorical}.

  \item Variable 10: \textbf{Capillary refill time} (< 3 seconds; >= 3 seconds)\\
    Comments: could have been a continuous variable originally.\\
    Decision: \texttt{Categorical}.

  \item Variable 11: \textbf{Pain} (alert, no pain; depressed; intermittent mild pain; intermittent severe pain; continuous severe pain)\\
   Comments: a subjective judgment of the horse's pain level. Despite donor's advice, these values are clearly ordered, so we consider it to be an ordinal variable ('the more painful, the more likely it is to require
            surgery').\\ %This decision could be revised.\\
    Decision: \texttt{Ordinal}.

  \item Variable 12: \textbf{Peristalsis} (hypermotile; normal; hypomotile; absent)\\ 
   Comments: an indication of the activity in the horse's gut (note order has to be reversed).\\
    Decision: \texttt{Ordinal}.

  \item Variable 13: \textbf{Abdominal distension} (none; slight; moderate; severe)\\
   Comments: none.\\
    Decision: \texttt{Ordinal}.
    
  \item Variable 14: \textbf{Nasogastric tube} (none; slight; significant)\\
   Comments: it refers to any gas coming out of the tube.\\
    Decision: \texttt{Ordinal}.

  \item Variable 15: \textbf{Nasogastric reflux} (none; <1 liter; >1 liter)\\
  Comments: none.\\
    Decision: \texttt{Ordinal}.

  \item Variable 16: \textbf{Nasogastric reflux PH}. 
  Comments: none.\\
    Decision: \texttt{Continuous}.
    
  \item Variable 17: \textbf{Rectal examination - feces} (normal; increased; decreased; absent)\\
   Comments: The values are re-ordered as: absent; decreased; normal; increased. Could it be considered categorical?\\
    Decision: \texttt{Ordinal}.

  \item Variable 18: \textbf{Abdomen} (normal; firm feces in the large intestine; distended small intestine; distended large intestine; other)\\
  Comments: none.\\
    Decision: \texttt{Categorical}.

  \item Variable 19: \textbf{Packed cell volume}\\
  Comments: the number of red cells by volume in the blood.\\
    Decision: \texttt{Continuous}.
  
  \item Variable 20: \textbf{Total protein}\\ 
    Comments: none.\\
    Decision: \texttt{Continuous}.

  \item Variable 21: \textbf{Abdominocentesis appearance} (clear; cloudy; serosanguinous)\\
    Comments: appearance of fluid obtained from the abdominal cavity.\\
    Decision: \texttt{Categorical}.
    
  \item Variable 22: \textbf{Abdominocentesis total protein}\\
    Comments: none.\\
    Decision: \texttt{Continuous}.
    
\end{itemize}
}

After this process, the dataset is described by 21 variables: 6 categorical, 7 continuous and 8 ordinal.

\subsection{Experimental settings}
The SNN is compared to two RBFs, as described next:

\begin{description}
	\item[RBFk:] a standard RBF where the centers are decided using the $k$-means clustering algorithm, and the hidden-to-output weights are set by solving the regularized least squares problem. The value of the smoothing parameter $\sigma^2$ is set according to the method described in \cite{Bishop95}.
	\item[RBF2:] a standard RBF where the centers are decided using the \textsc{Leader 2} clustering algorithm, and the hidden-to-output weights are set by solving the regularized least squares problem. The value of the smoothing parameter is set in the same way than for the RBFk.
%	\item[SVM-C:] a standard SVM with the RBF kernel for classification.
\end{description}

The reason for choosing two RBFs instead of one lies in the interest
in assessing any differences due to the clustering algorithm, since
the \textsc{Leader 2} algorithm can also be used to set the centers of
a standard \emph{RBF} network. This way it is easier to separate the
effect of the similarity processing. The two RBFs need a
pre-processing of the data, carried out following the recommendations
in \cite{Proben}. The input variables for the RBFs are then
standardized (to zero mean, unit standard deviation). This is not
needed by the SNN, but is beneficial for the RBF methods.  The values
of $\sminL$ for the SNN and the RBF2 as well as the value for $k$ in
$k$-means for the RBFk are chosen after some preliminary trials.

The resampling method used in this work is based on Dietterich
\cite{Dietterich98}. This method consists in five repetitions of
two-fold cross-validation (5$\times$2 CV), returning ten test set
performance estimations, that are combined as:

\begin{equation}
  t = \frac{p_{1}^{(1)}}{\sqrt{\frac{1}{5}\sum_{i=1}^5 s_{i}^2 }}
%\label{functttest}
\end{equation}
\noindent

where $p_{i}^{(j)} = {p_{i}^{(j)}}[A] - {p_{i}^{(j)}}[B]$ is the
difference between the proportions of the two methods ($A, B$) being
compared, in partition ($i$) of replication ($j$), for $i \in
\{1,\dots,5\}$ and $j \in \{1, 2\}$; then we have the estimated
variances $s_{i}^2 = \left( p_{i}^{(1)} - \overline{p_i}\right)^2 +
\left( p_{i}^{(2)} - \overline{p_i}\right)^2$, where $\overline{p_i} =
(p_{i}^{(1)}+p_{i}^{(2)})/{2}$.

A paired $t$ test can then be computed to assess statistical
significance in any possible differences in performance.  The
hypothesis of both methods having the same error rate can be rejected
at the $95\%$ level when $t > 2.571$.

%Attending to the configuration of the experimental settings, they chose 
%2-fold cross validation because it gives large test sets and disjoint 
%training sets. They need a big test because they only use a single 
%difference $p_{1}^{(1)}$ in the test. The disjoint training sets give 
%some independence, situation assumed that is not real because in the 
%replications all the instances are reused.\\
%They chose 5 replications of the cross validation because using 5 or 
%more replication increases the risk of Type-I error, and that is which 
%we want to exploit.\\
%They use only one of the observed differences of proportions rather than 
%the mean of all of them because this mean tends to overestimate the true 
%difference, attending to their reasoning. This is a problem generated by 
%the lack of independence between the folds of the cross validation. That 
%would justify the use of only one difference, so they had to make the 
%assumption that the variance estimates $s_i$ are independent of $p_{1}^{(1)}$ 
%and that $p_{1}^{(1)}$ is the better difference $p_{i}^{(j)}$.

As an alternative, Alpaydin defends the use of an $F$ test, where all
the differences are combined \cite{Alpaydin99}:

\begin{equation}
  F = \frac{\sum_{i=1}^5 \sum_{j=1}^2 \left( p_{i}^{(j)}\right)^2}{2\sum_{i=1}^5 s_{i}^2 }
%\label{functttest}
\end{equation}

%He defines a statistic computed from the errors on the test set of the 
%two methods adjusted with the training set. They follow the null 
%hypothesis, that is, both methods have the same error rate. If this 
%assumption holds, the statistic obeys a certain distribution.\\
%In this situation, only two kind of errors could happen. If we reject 
%the hypothesis when no difference between the errors exists, we incur 
%a type-I error. Otherwise, if we accept the null hypothesis when a 
%difference exists, we incur a type-II error. Using this last error, he 
%defines the \emph{power} of the test ($1 - Pr[Type-II error]$) as the 
%probability of detecting a difference when a difference exists.

The hypothesis of both methods having the same error rate can be
rejected at the $95\%$ level when $F > 4.74$.  This approach combines
better the ten statistics calculated and thus can be
expected to increase the robustness at no additional cost.

\subsection{Discussion}
\label{sec:Discussion}

The prediction errors for the \textsc{HC23} problem are displayed in
Table \ref{efHC23}. The three results are very similar, with a slight
advantage for the SNN, that could be perfectly due to sampling
variability. This is confirmed by the obtained statistical
significances, displayed in Table \ref{efHC23tests}. In this table,
performance of the SNN is compared to that of the two RBFs. The three
hypotheses raised are that SNN performance is not equal to the other
two, one by one. Only the $t$ test for averaged classification error
turns out to be significant (though by a small margin). The other
hypotheses cannot be rejected (also by a small margin).
The conclusion is that none of the networks is able to adequately
capture the complex relationship between predictors and target class,
attending to the relatively high normalized squared errors.

\begin{table}[!hbt]
\caption{
  \label{efHC23}
  Averaged prediction errors for the \textsc{HC23}
  problem. \emph{Error} is the percentage of errors; \emph{MSE} is the
  mean squared error; \emph{NRMSE} is the normalized MSE.
}
\begin{center}
\begin{tabular}{l|rc|rc|rc}
Method & \multicolumn{2}{c|}{\emph{Error}} & \multicolumn{2}{c|}{\emph{MSE}} & \multicolumn{2}{c}{\emph{NRMSE}} \\
\hline
SNN      & 32.79 &  & 0.147 &  & 0.901 &  \\
\emph{RBF2} & 33.33 &  & 0.149 &  & 0.907 &  \\
\emph{RBFk} & 33.22 &  & 0.148 &  & 0.906 &  \\
%\emph{SVM}  & 65.25 &  &     & | &     & | \\
\end{tabular}
\end{center}
\end{table}

\begin{table}[!hbt]
\caption{
  \label{efHC23tests}
  Statistical significances for the \textsc{HC23} problem. Positive
  results are shown in bold.
}
\begin{center}
\begin{tabular}{l|rrr}
Obs. & Test & \emph{RBF2} & \emph{RBFk} \\
\hline
\emph{Error} & $t$ & \textbf{2.692} &  2.050\\
           & $F$ & 1.783 & 1.193 \\
\hline
\emph{MSE}  & $t$ &  1.880 & 2.251 \\
           & $F$ & 0.847 & 1.058 \\
\hline
\emph{NRMSE}  & $t$ & 1.894 &  2.26 \\
           & $F$ & 0.855 & 1.056 \\
\hline
\end{tabular}
\end{center}
\end{table}

% \begin{table}[!hbt]
% \caption{
%   \label{efHC23tests}
%   Statistical significances for the \textsc{HC23} problem.
% }
% \begin{center}
% \begin{tabular}{l|rrrr}
% Obs. & Test & \emph{RBF2} & \emph{RBFk} & \emph{SVM} \\
% \hline
% \emph{Error} & $t$ & \textbf{2.692} &  2.05 & 0.755\\
%            & $f$ & 1.783 & 1.193 &  2.38\\
% \hline
% \emph{MSE}  & $t$ &  1.88 & 2.251 &    --\\
%            & $f$ & 0.847 & 1.058 &    --\\
% \hline
% \emph{NRMSE}  & $t$ & 1.894 &  2.26 &    --\\
%            & $f$ & 0.855 & 1.056 &    --\\
% \hline
% \end{tabular}
% \end{center}
% \end{table}

The prediction errors for the \textsc{HC24} problem are displayed in
Table \ref{efHC24}. This time SNN performance seems much better than that
of the two RBFs. This is also suggested by the obtained statistical
significances, displayed in Table \ref{efHC24tests}. Rather
surprisingly, the $t$ and $F$ tests do not turn out to be significant
for averaged classification error, but they are for the two types of
squared errors. This effect can be caused by the rather indirect
relation between classification accuracy and squared error. Given that
the networks were trained to minimize squared errors, these should be
the quantities to be taken as reference to evaluate performance.

A different matter is overfitting analysis. If a network obtains lower
squared errors because it concentrates on reducing the error on
certain examples (and ignoring others), this would be reflected in a
similar or worse classification accuracy. Since this is not the case
(lower square errors entail lower prediction errors) the conclusion is
that the SNN shows a better fit for the problem. This said, the
relatively high normalized squared errors point out that the fit could
be much better.

\begin{table}[!hbt]
\caption{
  \label{efHC24}
 Averaged prediction errors for the \textsc{HC24}
  problem. \emph{Error} is the percentage of errors; \emph{MSE} is the
  mean squared error; \emph{NRMSE} is the normalized MSE.
}
\begin{center}
\begin{tabular}{l|rc|rc|rc}
Method & \multicolumn{2}{c|}{\emph{Error}} & \multicolumn{2}{c|}{\emph{MSE}} & \multicolumn{2}{c}{\emph{NRMSE}} \\
\hline
SNN      & 16.73 &  & 0.128 &  &  0.740 \\
\emph{RBF2} & 20.01 &  & 0.152 &  & 0.808 \\
\emph{RBFk} & 20.06 &  & 0.153 &  & 0.809 \\
%\emph{SVM}  & 80.056 &  &     & | &     & | \\
\end{tabular}
\end{center}
\end{table}

\begin{table}[!hbt]
\caption{
  \label{efHC24tests}
  Statistical significance for the \textsc{HC24} problem. Positive
  results are shown in bold.
}
\begin{center}
\begin{tabular}{l|rrr}
Obs. & Test & \emph{RBF2} & \emph{RBFk}\\
\hline
\emph{Error} & $t$ & 0.505 &  1.370 \\
           & $F$ & 1.874 & 3.053 \\
\hline
\emph{MSE}  & $t$ & 2.286 & \textbf{2.986}  \\
           & $F$ & \textbf{15.119} & \textbf{12.786}  \\
\hline
\emph{NRMSE}  & $t$ & 2.351 & \textbf{2.755}  \\
           & $F$ & \textbf{14.602} & \textbf{10.346} \\
\hline
\end{tabular}
\end{center}
\end{table}

% \begin{table}[!hbt]
% \caption{
%   \label{efHC24tests}
% Statistical significance for the \textsc{HC24} problem.
% }
% \begin{center}
% \begin{tabular}{l|rrrr}
% Obs. & Test & \emph{RBF2} & \emph{RBFk} & \emph{SVM} \\
% \hline
% \emph{Error} & $t$ & 0.505 &  1.37 &  0.22\\
%            & $f$ & 1.874 & 3.053 & 2.584\\
% \hline
% \emph{MSE}  & $t$ & 2.286 & \textbf{2.986} &    \\
%            & $f$ & \textbf{15.119} & \textbf{12.786} &    \\
% \hline
% \emph{NRMSE}  & $t$ & 2.351 & \textbf{2.755} &    \\
%            & $f$ & \textbf{14.602} & \textbf{10.346} &    \\
% \hline
% \end{tabular}
% \end{center}
% \end{table}

A final point is made on the causes of this differential performance
among the nets. The only difference in the two RBFs is found in the
way the first-layer weights (the centers) are set up. The virtually
non-existent differences among the two methods (for both problems)
suggest that the clustering algorithm is playing no role in SNN
performance. Therefore, any difference should be attributable to the
similarity processing.

\section{Conclusions}
\label{sec:Conclusions}

In a neural network training process, the hidden layer(s) try to find
a new, more convenient representation for the problem {\it given} the
data representation chosen, a crucial factor for a
successful learning process. This part of the solution can be seen as
a clever pre-processing of the dataset to better capture the
underlying similarity relations between the examples. Subsequent
(similarity) data processing can be delivered to linear modeling
methods (such as those used in this paper). However, nothing prevents
the use of a non-linear method (as a support vector machine) that will
hopefully solve better the remaining target variability.

We advocate for the use of expert knowledge, whenever available, to
choose the best similarity functions for each variable. This part is a
two-blade sword, in that the enormous flexibility of the method is in
balance to the laborious work of analysing each variable and taking
appropriate decisions.  Since we are not able to contemplate all data
types in advance, we have presented a set of very basic partial
similarity functions that can be taken as \emph{default} methods.

The interpretability of the network is greatly enhanced, for two reasons. First, the neurons are centered at known examples acting as prototypes. Second, the output of the $k$-th neuron is a linear combination of the similarities of the input to the set of prototype neurons.

We reckon that much work remains to be done until the SNN can turn into a viable off-the-shelf modeling method. In this sense, and in the light of the analysed Horse Colic problem, it is worth noticing that the SNN could be further taylored with ease, something much more difficult in the case of the two RBFs.

\end{document}